\documentclass[10pt,twocolumn,letterpaper]{article}

\usepackage{3dv}
\usepackage{times}
\usepackage{epsfig}
\usepackage{graphicx}
\usepackage{amsmath}
\usepackage{amssymb}
\usepackage{multirow}
\usepackage{threeparttable}

\usepackage[pagebackref=true,breaklinks=true,colorlinks,bookmarks=false]{hyperref}

\threedvfinalcopy 

\def\threedvPaperID{16} 

\ifthreedvfinal\fi
\begin{document}

\title{ORFD: A Dataset and Benchmark for Off-Road Freespace Detection}

\author{Chen Min$^{1,2}$ and Weizhong Jiang$^{2}$ and Dawei Zhao$^{2,*}$ and Jiaolong Xu$^{2}$ \\	
	and Liang Xiao$^{2}$ and Yiming Nie$^{2}$ and Bin Dai$^{2}$\\
$^1$Peking University\\
$^2$NIIDT\\
{\tt\small minchen@stu.pku.edu.cn}
}

\maketitle

\begin{abstract}
   Freespace detection is an essential component of autonomous driving technology and plays an important role in trajectory planning. In the last decade, deep learning based freespace detection methods have been proved feasible. However, these efforts were focused on urban road environments and few deep learning based methods were specifically designed for off-road freespace detection due to the lack of off-road dataset and benchmark. In this paper, we present the ORFD dataset, which, to our knowledge, is the first off-road freespace detection dataset. The dataset was collected in different scenes (woodland, farmland, grassland and countryside), different weather conditions (sunny, rainy, foggy and snowy) and different light conditions (bright light, daylight, twilight, darkness), which totally contains 12,198 LiDAR point cloud and RGB image pairs with the traversable area, non-traversable area and unreachable area annotated in detail. We propose a novel network named OFF-Net, which unifies Transformer architecture to aggregate local and global information, to meet the requirement of large receptive fields for freespace detection task. We also propose the cross-attention to dynamically fuse LiDAR and RGB image information for accurate off-road freespace detection. Dataset and code are publicly available at \url{https://github.com/chaytonmin/OFF-Net}.
\end{abstract}

\section{Introduction}
\begin{figure}
	\centering
	\begin{minipage}[b]{.8\linewidth}
		\centerline{\includegraphics[width=8cm]{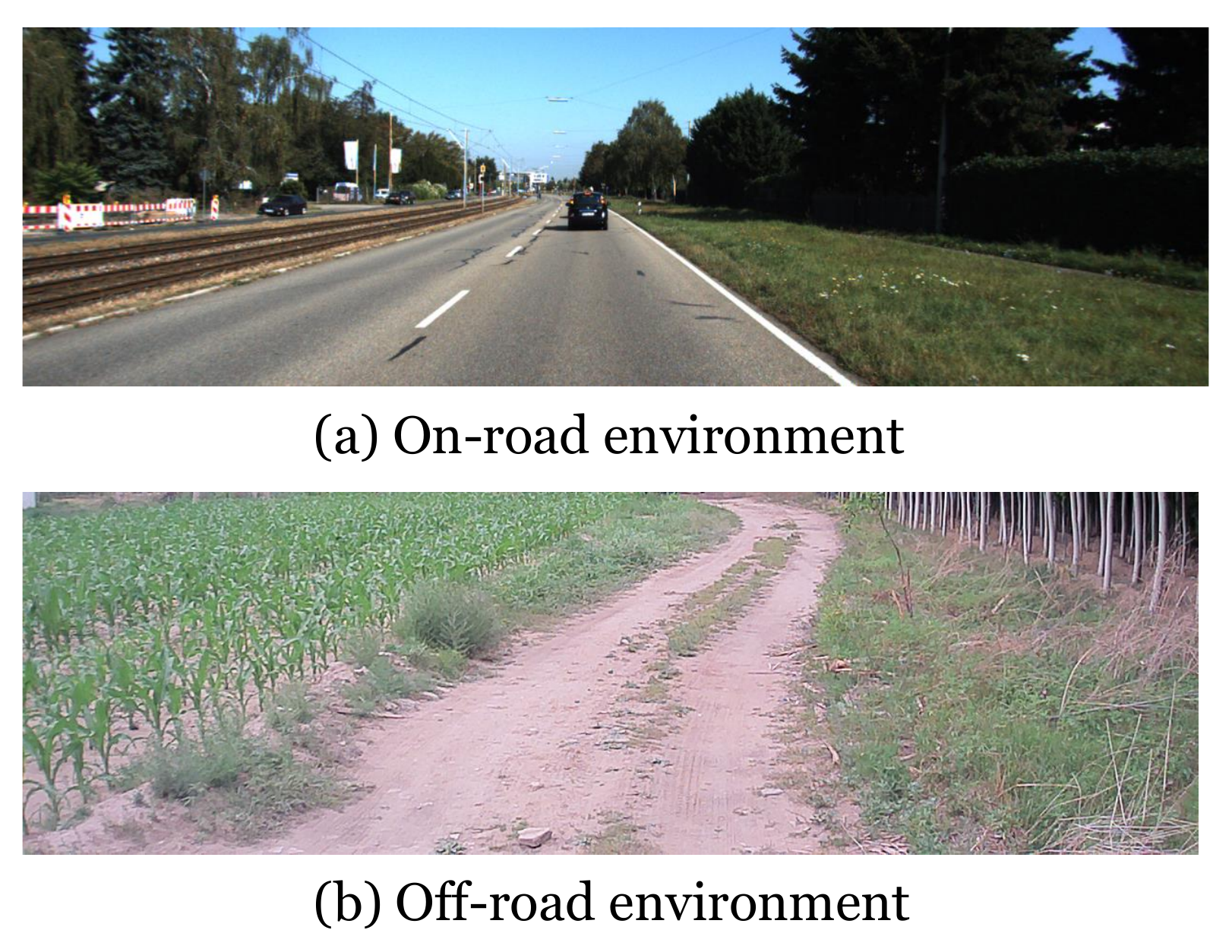}}
	\end{minipage}
	\caption{Difference between the on-road environment ({\ie}, the urban environment) and the off-road environment. The on-road environment provides many structural cues, such as lane markings and roadway signs while there are no well paved and clearly outlined roadways in off-road environment.}
	\label{difference}
\end{figure}
The ultimate goal of the development of autonomous driving is to liberate the driver, that is, no human attention is needed for any road environment. In order to achieve this goal, it is necessary to study various road environments, not limited in on-road environments, while the existing researches are mostly focused on on-road environments. Freespace detection, one of the most important technology of autonomous driving, has different concepts for structured on-road and unstructured off-road scenes, as shown in Fig.~\ref{difference}. For the former, freespace mainly refers to regular roads, while for off-road scenes, the concept of freespace is relatively vague. The autonomous vehicles need to pass through grassy, sandy or muddy off-road environments. This is a big challenge for autonomous vehicles since the off-road environments are complex and diverse, for example, tall grass and short grass are very different in terms of traversability, because tall grass may hide invisible obstacles or holes. To our knowledge, there are relatively few researches on the detection of freespace in off-road  environments~\cite{rugd}.

Data-driven based methods have achieved great success in the last decade and the world has entered an era of deep learning. With the help of data-driven based deep learning methods, many problems of autonomous driving have been solved and autonomous vehicle is becoming a reality. To improve the performance of deep learning methods for autonomous driving, a great many of datasets for scene perception of autonomous driving have been published, such as KITTI~\cite{kitti}, nuScenes~\cite{nuscenes}, Waymo~\cite{waymo} and so on. However, the existing published autonomous driving datasets are mainly collected in the urban cities, since the mainstream self-driving companies focus on autonomous driving technology in urban environments. Very few datasets are collected in off-road environments~\cite{rellis-3d}. 

Freespace detection, also known as traversable area detection, is an essential component of the autonomous driving technology and plays an important role in path planning both in on-road and off-road environments. However, the existing off-road datasets~\cite{deepscene, ycor,rugd,rellis-3d} do not focus on the traversability analysis in off-road environments. As such, there is a need for a dataset focused on freespace detection task in off-road environments. To address this, we present the \textbf{OFF-Road Freespace Detection} (\textbf{ORFD}) dataset, which was collected from the off-road scenes for promoting deep learning research in off-road environments. Taking into account the complexity and diversity of off-road environments and the influence of different light factors and seasonal factors on autonomous driving, we collected the ORFD dataset in different seasons (spring, summer, autumn and winter), at different time (day, evening and night) and in different scenes (such as woodland, farmland, grassland and countryside). The dataset includes a total of $12,198$ LiDAR point cloud and RGB image pairs. We synchronized the LiDAR and RGB image data and labeled the freespace in the image plane. Referring to the processing method of the KITTI urban road dataset~\cite{kitti_road}, we projected the LiDAR point cloud onto the corresponding RGB image plane, so as to obtain the depth information of the freespace.

We also introduce an off-road freespace detection benchmark, called OFF-Net. A Transformer architecture is utilized to enlarge the receptive field and capture the context information. Similar to on-road freespace detection task, one modality is not enough for accurate freespace detection. We propose a cross-attention mechanism, that outputs the modality weights to help dynamically fuse LiDAR point cloud and RGB image information. Experiments show that our algorithm achieves good performance on the off-road freespace detection task evaluated on the ORFD dataset.

The highlights of our work are as follows:
\begin{itemize}
	\item We propose the first off-road freespace detection dataset, called ORFD dataset, which covers different off-road scenes (woodland, farmland, grassland and countryside), different weather conditions (sunny, rainy, foggy and snowy) and different light conditions (bright light, daylight, twilight, darkness) for the generalization of off-road freespace detection.
	\item We propose OFF-Net, a Transformer network architecture, to aggregate the context information for accurate off-road freespace detection. We also introduce the cross-attention mechanism to dynamically fuse data from both camera and LiDAR to leverage the strengths of each modality.
	\item Our work with the dataset and benchmark will open new research area of the off-road freespace detection to ultimately enhance the perception ability of the autonomous vehicle in off-road environments.
\end{itemize}

\section{Related Work}

\subsection{Datasets}

Autonomous driving technology has developed rapidly in recent years with the help of deep learning methods. The success of deep learning relies heavily on the large-scale training data. According to different collection environments, road dataset for autonomous driving can be divided into two classes: on-road datasets and off-road datasets. The on-road datasets, such as KITTI~\cite{kitti}, SemanticKITTI~\cite{semantickitti}, nuScenes~\cite{nuscenes} and Waymo~\cite{waymo} have been widely used for autonomous driving. However, there are few datasets for off-road environments, and now, we will review the existing off-road datasets.

To our knowledge, DeepScene~\cite{deepscene} is the first public off-road dataset for multispectral segmentation collected in unstructured forest environments. YCOR~\cite{ycor} is a more diverse and challenging dataset than DeepScene. However, both DeepScene and YCOR have small amount of data. RUGD~\cite{rugd} was collected in an off-road environment for RGB image semantic segmentation task but only has one modality data. RELLIS-3D Dataset~\cite{rellis-3d} improves RUGD with multi-modal sensor to enhance autonomous navigation in off-road environments. The above datasets do not focus on the analysis of traversability in off-road environments, while the detection of traversable area, namely freespace, is one of the most important modules for autonomous driving, especially in off-road environments. In our work, we create the first dataset with accurate and complete ground truths for off-road freespace detection.

\subsection{Methods}

The existing freespace detection methods are mostly designed for on-road navigation and most of them fuse the LiDAR and camera information together for accurate freespace detection. RBNet~\cite{rbnet} detects both road and road boundary in a single process and models them with a Bayesian network. TVFNet~\cite{tvfnet} takes the LiDAR imageries and the camera-perspective maps as inputs and outputs pixel-wise road detection results in both the LiDAR's imagery view and the camera's perspective view simultaneously. LC-CRF~\cite{lc-crf} detects road with LiDAR-camera fusion in a conditional random field (CRF) framework to exploit both range and color information.  RBANet~\cite{rbanet} proposes the reverse attention and boundary attention units for road segmentation. PLARD~\cite{plard} introduces a progressive LiDAR adaptation-aided road detection approach to adapt LiDAR information into visual image-based road detection and improve detection performance. SNE-RoadSeg~\cite{sne} infers surface normal information from dense depth images and fuses it with image information to improve the freespace detection performance. The above methods are focused on the on-road freespace detection. In this paper, we introduce a novel Transformer architecture based method for off-road freespace detection and do extensive experiments on our off-road dataset.

\section{ORFD Dataset}
\begin{figure}[t]
	\centering
	\centerline{\includegraphics[width=3.4in]{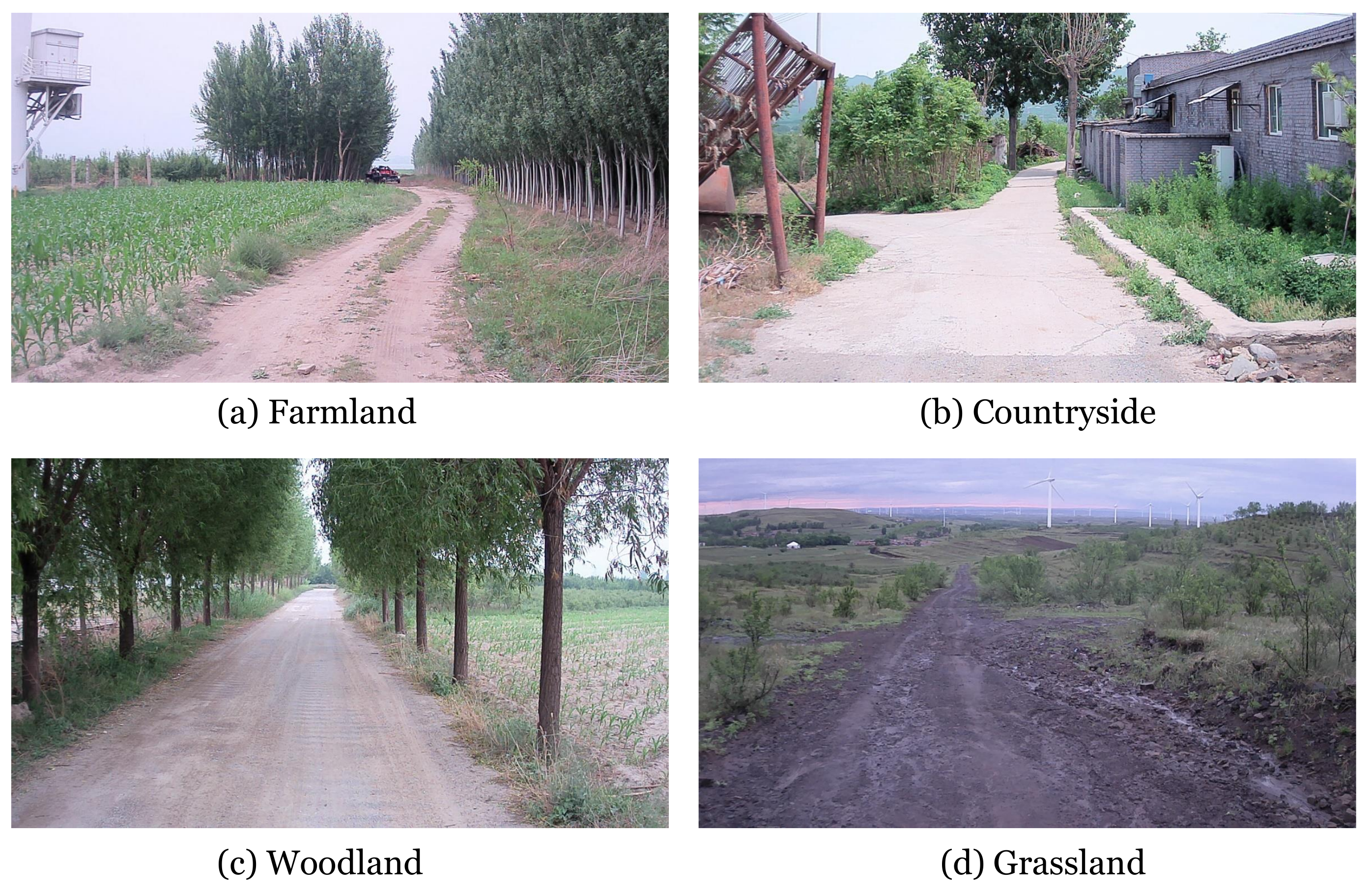}}
	\caption{ORFD dataset contains a variety of scenes for the generalization of off-road freespace detection.}
	\label{type}
\end{figure}

\begin{figure}[t]
	\centering
	\centerline{\includegraphics[width=3.4in]{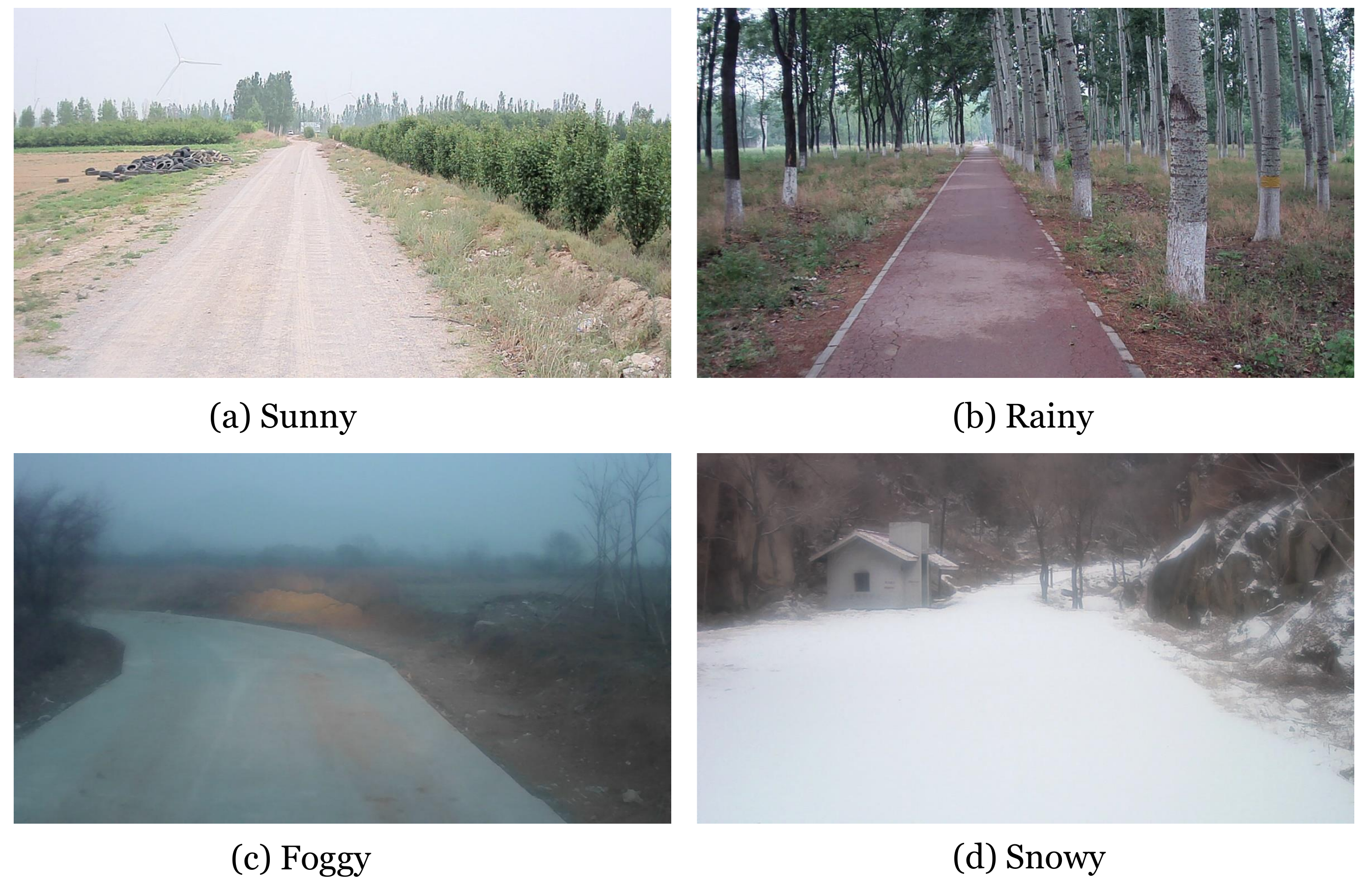}}
	\caption{Different weather conditions are considered in ORFD dataset.}
	\label{weather}
\end{figure}

\begin{figure}[t]
	\centering
	\centerline{\includegraphics[width=3.4in,height=2.2in]{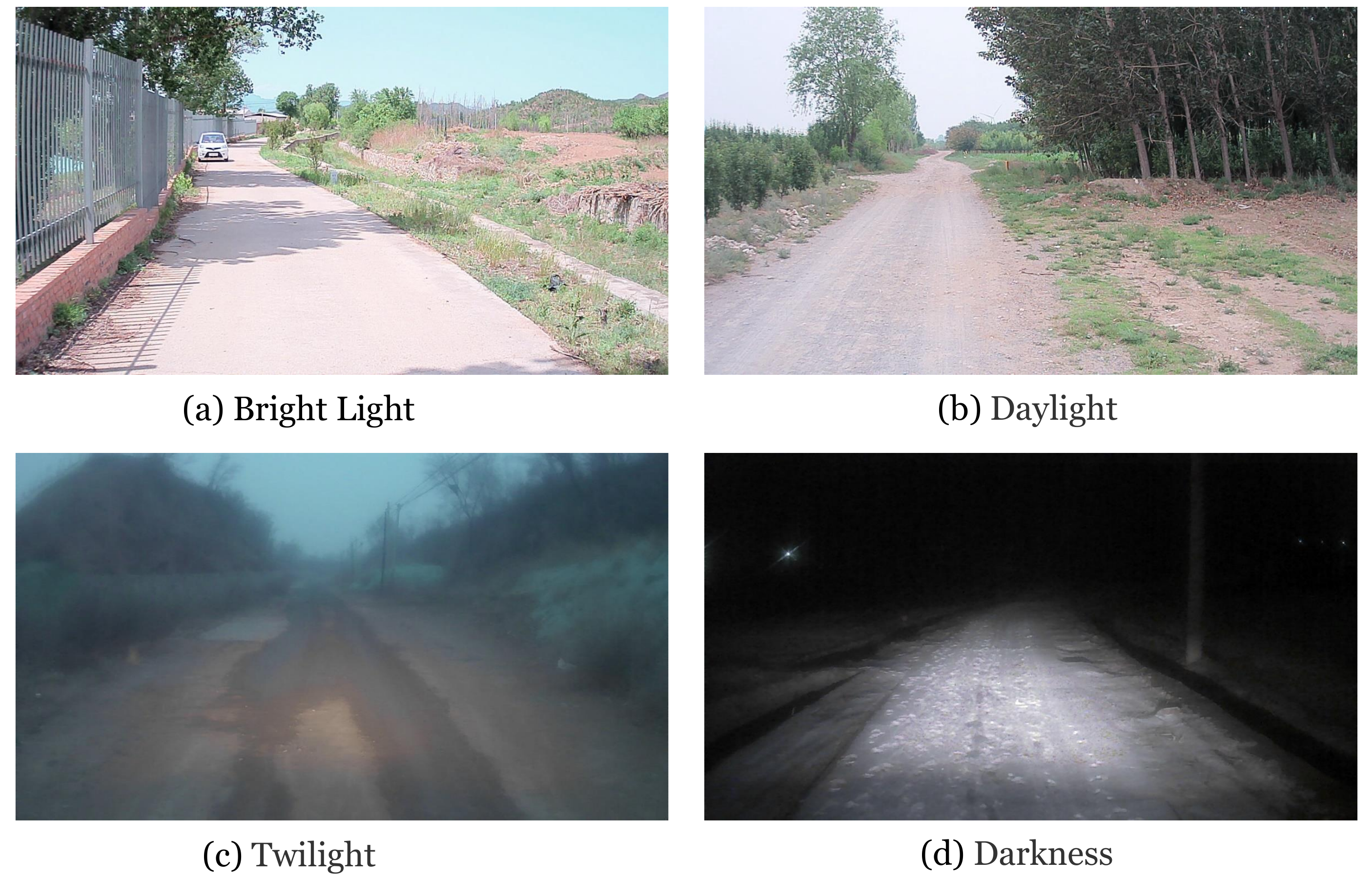}}
	\caption{ORFD dataset was collected at different time of the day to cover the light conditions affecting the autonomous navigation.}
	\label{day}
\end{figure}

\begin{figure}[t]
	\centering
	\centerline{\includegraphics[width=3.4in,height=2.218in]{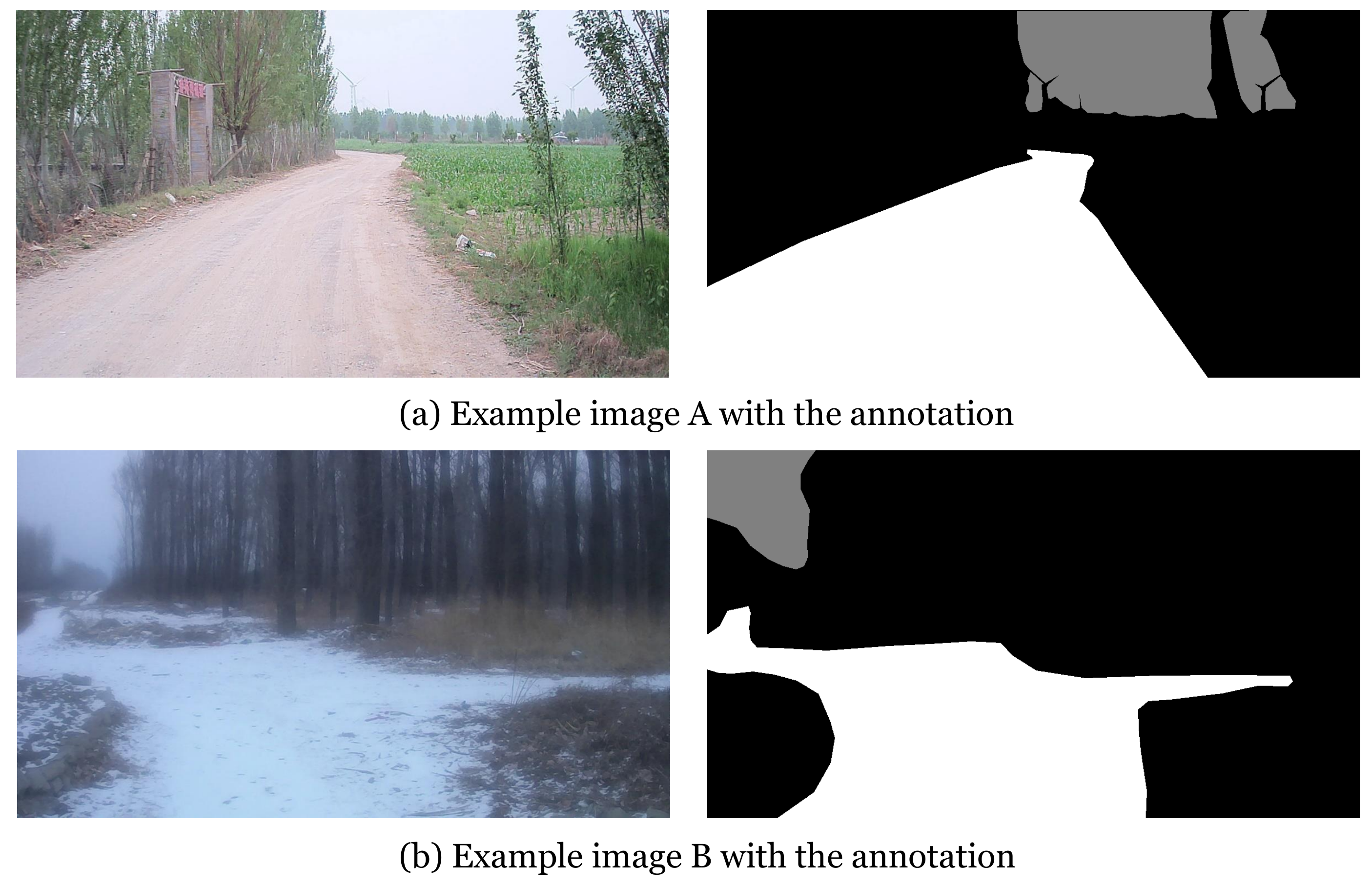}}
	\caption{We adopt the concept of 'freespace' to label the ORFD dataset and provide the annotation of three classes, {\ie}, traversable area (in white color), non-traversable area (in black color) and unreachable area (in gray color).}
	\label{annotation}
\end{figure}
This section describes the dataset we created for off-road freespace detection called the ORFD dataset. The dataset containing LiDAR point cloud and RGB image information, was collected in a variety of off-road scenes for facilitating deep learning research in off-road environments.

\subsection{Data Description}\label{data-description}

ORFD dataset was collected in off-road environments. Compared with the structured on-road environments, off-road environments vary greatly due to terrain, vegetation, season, weather, time and so on. Therefore, we collected the off-road dataset including a variety of scenes (such as woodland, farmland, grassland and countryside) as shown in Fig.~\ref{type}, different season and weather conditions (such as sunny, rainy, foggy and snowy weather from spring to winter) as shown in Fig.~\ref{weather}, and different light conditions (bright light, daylight, twilight, darkness) as shown in Fig.~\ref{day}. The statistics are shown in the Table~\ref{split}. We collected $30$ sequences in various off-road environments in China, and one sequence covers a distance of about $100$ meters. We annotated a total of $12,198$ LiDAR point cloud and RGB image pairs. The LiDAR is $40$-line, and the size of RGB image is $1280 \times 720$.

\subsection{Data Annotations}

We provide pixel-wise image annotations of ORFD dataset for off-road freespace detection. There are three classes: traversable area, non-traversable area and unreachable area as shown in Fig.~\ref{annotation}. As mentioned earlier, in off-road environments we use the concept of freespace instead of the road, because the road in the usual on-road environment is clear shown in Fig.~\ref{difference} (a), while there may not be such a regular road in off-road environments, as shown in Fig.~\ref{difference} (b), where autonomous vehicles can indeed move in. Therefore, we believe that in the off-road scenes, areas that do not pose a threat to the safety of the autonomous vehicles can be regarded as freespace or traversable area. The non-traversable area is mainly an area composed of objects in the scene that pose a threat to the safe driving of autonomous vehicles. Unreachable area mainly refers to the area composed of objects that are relatively far away, which temporarily does not pose a threat to the safe driving of autonomous, and a typical example is the sky. Therefore, we mainly carry out pixel-level labeling of three types of objects: traversable area, non-traversable area and unreachable area, as shown in Fig.~\ref{annotation}.


\subsection{Dataset for Deep Learning}
\begin{figure}[t]
	\centering
	\centerline{\includegraphics[width=3.4in]{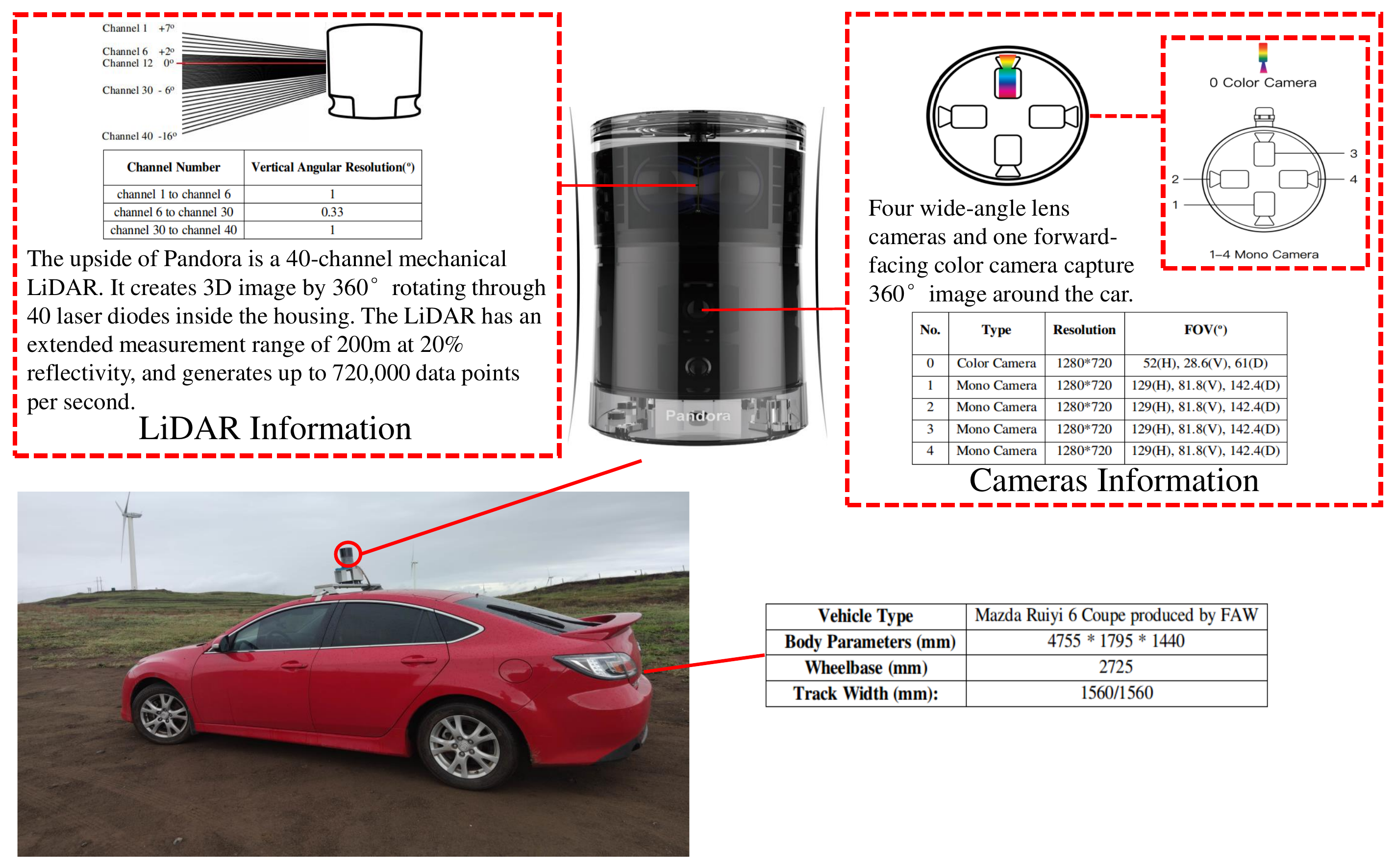}}
	\caption{Detail information of vehicle and sensor to collect LiDAR and camera data.}
	\label{sensor}
\end{figure}

Deep learning methods have shown promising results in on-road freespace detection with the public on-road datasets~\cite{kitti,semantickitti}. Only the RGB image lacking the geometry information is insufficient for freespace detection. We provide the ORFD dataset with the RGB image, LiDAR point cloud, calibration, sparse depth, dense depth and ground truth. The sparse depth was obtained by projecting the LiDAR point cloud to the image plane with the calibration information and then the sparse depth was interpolated to get the dense depth. We split the dataset into training, validation and testing set as shown in Table~\ref{split}. From the table, we can see that the ratio of the three sets is about $7:1:2$. As it is hard to define road in grassland, we only choose one sequence of well-annotated data in grassland for training set.

\begin{table*}[htbp]
	\caption{Training, validation and testing splits of the ORFD dataset.}
	\begin{center}
		\setlength{\tabcolsep}{0.1mm}{
			\begin{tabular}{c|cccc|cccc|cccc|c|c}
				\hline
				Split & Farmland &Woodland &Grassland & Countryside  &Sunny & Rainy & Foggy & Snowy&Bright
				light&Daylight & Twilight & Darkness & Total & $\%$  \\ 
				\hline
				Train& 2718 & 3180 &361&2139&3803 &2434&1136&1025 &1019 &4254 &927&2198&8398&68.8\\ 
				
				Val&356 &302 &0&587 &356 &302&0&587 &0 &356 &302&587&1245&10.2\\
				
				Test&1129 &1064 &0&362 & 1071 &405&720&359 &361 & 710 &359&1125&2555&20.9\\ 
				\hline 
				Total&4203 &4546 &361&3088 & 5230 &3141&1856&1971 &1380 & 5320 &1588&4510&12198&100\\ 
				\hline
			\end{tabular}
			\label{split}
		}
	\end{center}
\end{table*}

\subsection{Sensors}

The vehicle used to collect the ORFD dataset is the Mazda Ruiyi 6 Coupe produced by FAW, with the body parameters (mm): $4755 \times 1795 \times 1440$, and the wheelbase (mm): $2725$, and track width (mm): $1560/1560$. A sensor fusion kit Pandora produced by Hesai Technology is installed on the top of the vehicle to collect LiDAR point cloud and RGB image data. Pandora is composed of a $40$-line mechanical LiDAR at the upper part and $5$ cameras distributed around the lower part, including a color camera and $4$ wide-angle black-and-white cameras. Detailed parameters of vehicle and sensors are shown in Fig.~\ref{sensor}.

\subsection{Synchronization and Calibration}

Pandora controls the motor rotating and laser firing time of the LiDAR, and at the same time, LiDAR controls the exposure time and frame rate of the cameras. Therefore, Pandora can achieve synchronization of point cloud data from LiDAR and image data from the cameras. We use the method in~\cite{calib} to calibrate the external parameters of the LiDAR.

\section{Method}

\begin{figure}[t]
	\centering
	\begin{minipage}[b]{.8\linewidth}
		\centerline{\includegraphics[width=3.4in]{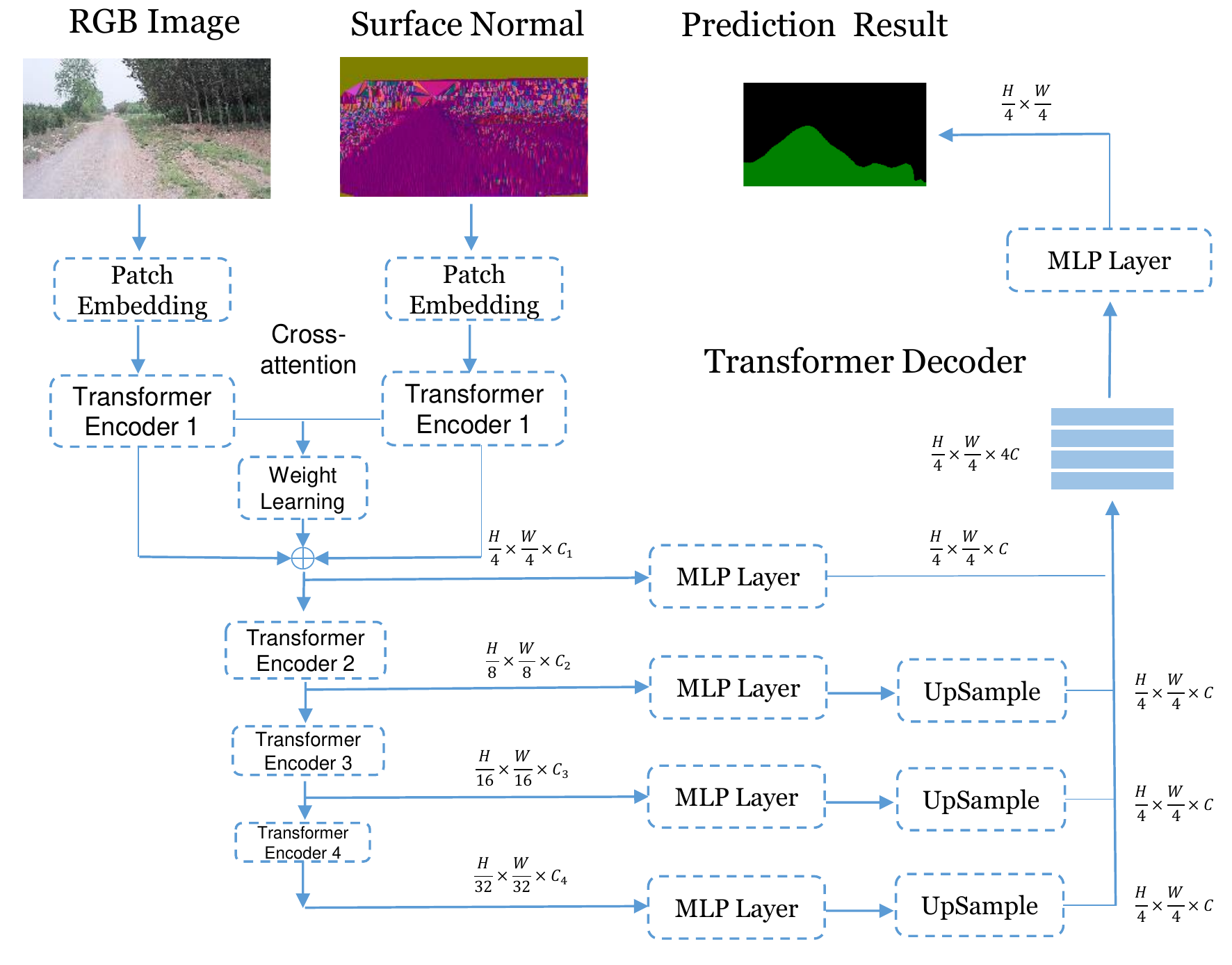}}
	\end{minipage}
	\caption{The architecture of our OFF-Net. The Transformer encoder extracts the features from both the RGB image and surface normal and a Transformer decoder predicts the freespace result. A cross-attention is designed to fuse data from both camera and LiDAR to dynamically leverage the strengths of each modality.}
	\label{flowchart}
\end{figure}

\subsection{Problem Definition}

We formulate the off-road freespace detection task as the pixel-wise classification problem on the RGB image plane, {\ie}, whether the pixel belongs to the traversable area. Given the RGB image $\mathbf{I}$ and LiDAR $\mathbf{L}$, the network F predicts the probability map. The goal of off-road freespace detection is to minimize the following loss function:
\begin{equation} \label{overall_loss_function}
\min_{\theta} \sum_{i}\mathcal{L} (F(\mathbf{I}_i,\mathbf{L}_i),\hat{\mathbf{Y}}_i),
\end{equation}
where $\theta$ is the parameter of network F and $\hat{\mathbf{Y}}$ is the freespace detection ground truth of the $i$-th training example.  

\subsection{OFF-Net}

We developed a network called OFF-Net to combine camera and LiDAR information ( {\ie}, surface normal information calculated from the LiDAR point cloud). The reason why we choose the surface normal information as the network input is that the points within the road have similar surface normals, and the surface normals are calculated from dense depth images using the method proposed by Fan~\cite{sn}.

As the freespace detection task needs the network to have large receptive field, however, the CNN has the limited receptive field. Inspired by the success of the Transformer framework in capturing local and global information. We introduce the Transformer network architecture proposed by Xie~\cite{segformer} into the off-road freespace detection task. The structure of the our OFF-Net is illustrated in Fig.~\ref{flowchart}, and the special modules are described as follows.

\subsubsection{Transformer Encoder}

In order to obtain multi-level features, we first perform patch embedding on RGB image and surface normal separately with the resolution of $H \times W \times 3$. The outputs of patch emmbedding are hierarchical feature maps and the corresponding spatial resolutions are $\lbrace1/4,1/8,1/16,1/32\rbrace$ of the input size. We then compute the multi-head self-attention function with the heads $\mathbf{Q}$, $\mathbf{K}$, $\textbf{V}$: 
\begin{equation} \label{self-attention}
Attention(\textbf{Q},\textbf{K},\textbf{V})=softmax(\frac{\textbf{Q}\textbf{K}^T}{\sqrt{d_{head}}})\textbf{V}.
\end{equation}
Since fixed position encoding will reduce the accuracy as described in~\cite{segformer}, we use $3 \times 3$ convolution to capture position information for the Transformer encoder. The core formula of position encoding is as follow:
\begin{equation} \label{mlp}
\textbf{x}_{out}=MLP(GELU (Conv_{3 \times 3}(MLP(\textbf{x}_{in}))))+\textbf{x}_{in},
\end{equation}
where $\textbf{x}_{in}$ is the feature from the multi-head self-attention part, $\emph{GELU}$ is the activation function proposed by Dan~\cite{gelu}, $\emph{MLP}$ is a fully connected neural network layer and $Conv_{3 \times 3}$ is a $3 \times 3$ convolutional layer.

LiDAR point cloud data contains spatial geometric information but lacks semantic information, while monocular RGB image contains higher-level semantic information of the environment but lacks structual information. And features from RGB image and surface normal contribute differently to the final result of the freespace detection. In order to obtain the best weights for these two types of modalities to improve the detection performance, we design a dynamic fusion module. This module first adds the RGB image features and the surface normal features from LiDAR point cloud, and then sends the superimposed result to the MLP layer with the sigmoid activation function to learn the cross-attention. The cross-attention mechanism is simple but efficient in learning the weight of each modality. In this way, after the processing of the dynamic fusion module, we can get the refined features. The calculation formulas are as follows:
\begin{equation} \label{cross-attention}
\begin{aligned}
Cross\_Attention=\sigma(\textbf{x}_{img\_in}+\textbf{x}_{sn\_in}),\\
\textbf{x}_{img\_out}=Cross\_Attention*\textbf{x}_{img\_in}+\textbf{x}_{img\_in},\\
\textbf{x}_{sn\_out}=(1-Cross\_Attention)*\textbf{x}_{sn\_in}+\textbf{x}_{sn\_in},
\end{aligned}
\end{equation}
where $\textbf{x}_{img\_in}$ and $\textbf{x}_{sn\_in}$ are the learned RGB image and surface normal features after the Transformer block, $\textbf{x}_{img\_out}$ and $\textbf{x}_{sn\_out}$ are the refined RGB image and surface normal features, and $\sigma$ is the sigmoid activation function. 

\subsubsection{Transformer Decoder}

The Transformer decoder is used to fuse local and global information, and it consists of only the MLP layer. The features from the Transformer encoder are first put into the MLP layer to aggregate channel-wise information, and then upsampled to the same size ({\ie}, 1/4 of the input size). Finally, the features are fused together to obtain the freespace detection result.

\subsubsection{Loss Function}
We use the binary cross-entropy loss for off-road freespace detection, defined as follow:
\begin{equation} \label{loss}
\begin{split}
loss = -\frac{1}{batch}\sum_{i=0}^{batch}\sum_{m=1}^{M}\sum_{n=1}^{N}\hat{\textbf{Y}}^{i}_{mn}log\textbf{O}^{i}_{mn},
\end{split}
\end{equation}
where $\textbf{O}^{i}_{mn}$ is the predicted probability of pixel $mn$ of the $i$-th training sample, and $\hat{\textbf{Y}}^{i}_{mn}$ is the corresponding ground truth.

\section{Experimental Results and Discussion}

In this section, a series of experiments are conducted to validate the performance of the proposed dataset and method.

\subsection{Experiments}
\subsubsection{Experimental Setting}

We evaluate the proposed OFF-Net method on our ORFD dataset with two classes ({\ie},
traversable area and non-traversable area, the unreachable area is regarded as non-traversable area for simplicity) and compare it with FuseNet~\cite{fusenet} and SNE-RoadSeg~\cite{sne}. FuseNet extracts features from RGB image and depth image and then fuses depth features into RGB feature maps with VGG16~\cite{vgg} network architecture, and SNE-RoadSeg extracts and fuses features from RGB image and surface normal information with ResNet-152~\cite{resnet} network architecture.

We use five common metrics for the performance evaluation of off-road freespace detection:  1) Accuracy=$\frac{TP+TN}{TP+TN+FP+FN}$, 2) Precision=$\frac{TP}{TP+FP}$, 3) Recall=$\frac{TP}{TP+FN}$, 4) F-score=$\frac{2{TP}^2}{2{TP}^2+TP(FP+FN)}$ and 5) IOU=$\frac{TP}{TP+FP+FN}$, where TP, TN, FP and FN represent the number of true positive, true negative, false positive, and false negative pixels, respectively.

\subsubsection{Implementation Details}

We use Pytorch~\cite{pytorch} to implement our method, and the model is trained with the momentum (SGDM)~\cite{sgdm} optimizer. The initial learning rate is $0.001$ and the batch size is set as $8$. All experiments are performed using $4$ Nvidia RTX 3090 GPU devices. The image size for training and testing is set as $1280 \times 704$. 

\subsection{Results and Discussions}
\begin{table*}[htbp]
	\caption{Quantitative results on the ORFD testing set}
	\begin{center}
		\setlength{\tabcolsep}{2.1mm}{
			\begin{tabular}{c|c|ccccc|cc}
				\hline
				Method & Modality &Accuracy & Precision & Recall & F-score & IOU&Params&Speed  \\ 
				\hline
				FuseNet~\cite{fusenet}&RGB + Sparse Depth&87.4\% & 74.5\% &85.2\%&79.5\%&66.0\%&50.0M&{\bf49 Hz}\\  
				SNE-RoadSeg~\cite{sne}&RGB + Surface Normal& 93.8\% & {\bf 86.7\%} &92.7\%&89.6\%&81.2\%&201.3M&12.5 Hz\\ 
				OFF-Net (ours)&RGB + Surface Normal& {\bf 94.5\%} & 86.6\% &{\bf 94.3\%}&{\bf 90.3\%}&{\bf82.3\%}&{\bf 25.2M}&33.9 Hz\\  
				\hline
			\end{tabular}
			\label{results}
		}
	\end{center}
\end{table*}
\subsubsection{Evaluation on ORFD Dataset}

Table~\ref{results} shows the quantitative results on the ORFD testing set. Compared to FuseNet with depth and RGB image information as input, SNE-RoadSeg performs much better, which is because that the freespace can be assumed as a ground plane on which the points have similar surface normals while have different depth, thus the surface normal is more suitable than depth for freespace detection task. Our OFF-Net outperfoms FuseNet by $10.8\%$ on F-score rate and $16.3\%$ on IOU rate. Compared to SNE-RoadSeg, our method obtains $0.7\%$  higher F-score rate and $1.1\%$ higher IOU rate. The results show that our method with the Transformer framework can capture more local and global information for accurate freespace detection performance. Our OFF-Net uses only $25.2$M parameters and takes $29.5$ ms per input, satisfying real-time requirement. It runs $7$ $\times$ smaller and $2.7$ $\times$ faster than SNE-RoadSeg.

\subsection{Ablation Studies}

\begin{figure*}[ht]
	\centering
	\begin{minipage}[b]{.8\linewidth}
		\centerline{\includegraphics[width=18cm]{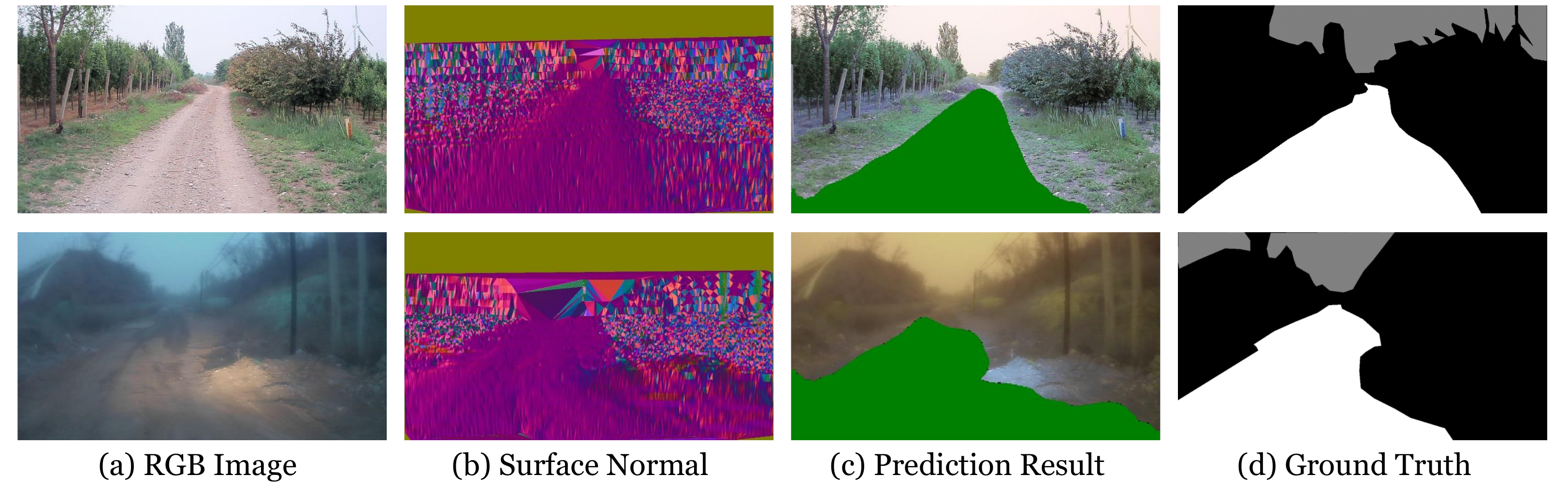}}
	\end{minipage}
	\caption{Qualitative results of our OFF-Net on the ORFD dataset. Our OFF-Net can predict off-road freespace accurately. There are also some unsatisfactory results as there are no concept of conventional roads.}
	\label{qualitative_result}
\end{figure*}

\begin{table}[htbp]
	\caption{Impacts of different model input.}
	\begin{center}
		\setlength{\tabcolsep}{0.01mm}{
			\begin{tabular}{c|ccccc}
				\hline
				Modality &Acc.& Pre.& Recall & F-score & IOU  \\ 
				\hline
				RGB & 88.8\% & 75.5\% &86.4\%&80.6\%&67.5\%\\
				Surface Normal & 93.3\% & 83.7\% &93.2\%&88.2\%&78.9\%\\
				RGB + Sparse Depth& 90.1\% & 76.7\% &90.9\%&83.2\%&71.3\%\\
				RGB + Dense Depth& 86.4\% & 68.4\% &92.3\%&78.6\%&64.8\%\\ 
				RGB + Surface Normal& {\bf94.5\%} &{\bf 86.6\%} &{\bf94.3\%}&{\bf90.3\%}&{\bf82.3\%}\\ 
				\hline
			\end{tabular}
			\label{different_input}
		}
	\end{center}
\end{table}

In this section, extensive ablation studies are performed to
validate several components in the ORFD dataset and the proposed OFF-Net method.

\subsubsection{Dataset}

To evaluate the influence of different part in the off-road DRFD dataset, we do experiments with different model input on the proposed OFF-Net. This ablation study, presented in Table~\ref{different_input}, shows that fusing two modalities can boost the freespace detection performance. The surface normal information is more important than depth information as the freespace has similar surface normals. But only using the surface normal information is not enough as the RGB image can provide the higher-level semantic information. It is needed to fuse the RGB image with the semantic information and LiDAR with the geometric information to leverage the strengths of each modality. Fusing the RGB image with the dense depth performs worse, as the dense depth was obtained by interpolating the sparse depth.

\subsubsection{Transformer Encoder}
We now analyze the influence of the number of transformer encoder block. In Table~\ref{ablation} we can observe that increasing the number of transformer encoder block can improve the performance with multi-level features. We use four transformer encoder blocks as the spatial resolution of stage 4 is very small. 

\subsubsection{Attention Mechanism}

In this section, we study the effect of the proposed cross-attention mechanism on the fusion of RGB image and surface normals from LiDAR point cloud. It can be seen from Table~\ref{ablation} that the cross-attention increases the IOU rate from $80.9\%$ to $82.3\%$, and the F-score rate from $89.4\%$ to $90.3\%$. Therefore, we can draw a conclusion that the proposed cross-attention mechanism can dynamically assign weights to the RGB image and surface normal information from LiDAR point cloud, and these weights are proven to be effective.

\begin{table}
	\caption{Impacts of Transformer encoder and cross-attention.}
	\begin{center}
		\setlength{\tabcolsep}{0.01mm}{
			\begin{tabular}{c|c|ccccc}
				\hline
				Encoder& Cross-attention &Acc. & Pre. & Recall & F-score & IOU  \\ 
				\hline
				1&$\checkmark$ & 92.5\% & 80.1\% &{\bf96.0\%}&87.3\%&77.5\%\\ 	
				2&$\checkmark$& 92.8\% & 81.1\% &95.6\%&87.7\%&78.2\%\\ 	
				3&$\checkmark$& 94.3\% & {\bf87.1\%} &92.7\%&89.8\%&81.5\%\\	
				4&$\checkmark$& {\bf94.5\%} & 86.6\% &94.3\%&{\bf90.3\%}&{\bf82.3\%}\\ 	
				4&$\times$& 94.1\% & 86.4\% &92.7\%&89.4\%&80.9\%\\
				\hline
			\end{tabular}
			\label{ablation}
		}
	\end{center}
\end{table}

\subsection{Qualitative Results}

It can be seen from Fig.~\ref{qualitative_result} that our OFF-Net can accurately estimate the off-road freespace. Specifically, OFF-Net has a good generalization ability for unseen scenes in the testing set. However, there are some failure cases, especially when the scene is very unstructured. Therefore, detecting freespace in an unstructured off-road environment is more challenging than in the on-road environment.

\section{Conclusion}
In this paper, we introduce an off-road freespace dataset, called the ORFD dataset, collected from a variety of off-road scenes, which, to our knowledge, is the first off-road freespace detection dataset. We believe that the ORFD dataset will help facilitate the study of autonomous navigation in off-road environments. We also introduce a new off-road freespace detection approach called OFF-Net, which unifies the Transformer network architecture to capture the context information. We design the cross-attention to dynamically aggregate information from both camera and LiDAR. In the future, we are planning to collect more freespace detection dataset to cover more off-road scenes.


{\small
\bibliographystyle{ieee_fullname}
\bibliography{egbib}
}

\end{document}